\documentclass[conference]{IEEEtran}
\IEEEoverridecommandlockouts

\usepackage{amsmath,graphicx}

\usepackage{subfigure} 
\graphicspath{./image/}

\usepackage{amssymb}
\usepackage{epstopdf}
\usepackage{tabularx}
\usepackage{multirow}
\usepackage{textcomp}
\usepackage{url}
\usepackage{lipsum}
\usepackage{makecell}

\usepackage{bm}
\usepackage{calc}

\usepackage{color}
\usepackage{dsfont}

\def\BibTeX{{\rm B\kern-.05em{\sc i\kern-.025em b}\kern-.08em
		T\kern-.1667em\lower.7ex\hbox{E}\kern-.125emX}}
\begin{document}
	
	\title{Deep Transfer Learning for Single-Channel Automatic Sleep Staging with Channel Mismatch
		\thanks{Corresponding author: \tt h.phan@kent.ac.uk.}
	}
	
	\author{\IEEEauthorblockN{Huy Phan$^{\dagger}$, Oliver Y. Ch\'{e}n$^{\ast}$, Philipp Koch$^{\ddagger}$, Alfred Mertins$^{\ddagger}$, and Maarten De Vos$^{\ast}$}\\
		\IEEEauthorblockA{\textit{$^{\dagger}$School of Computing, University of Kent, United Kingdom} \\
			\textit{$^{\ast}$Institute of Biomedical Engineering, University of Oxford, United Kingdom}\\
			\textit{$^\ddagger$Institute for Signal Processing, University of L\"ubeck, Germany}}
	}

	%
	
	\maketitle
	
	\begin{abstract}
		Many sleep studies suffer from the problem of insufficient data to fully utilize deep neural networks as different labs use different recordings set ups, leading to the need of training automated algorithms on rather small databases, whereas large annotated databases are around but cannot be directly included into these studies for data compensation due to channel mismatch.
		This work presents a deep transfer learning approach to overcome the channel mismatch problem and transfer knowledge from a large dataset to a small cohort to study automatic sleep staging with single-channel input. We employ the state-of-the-art \emph{SeqSleepNet} and train the network in the source domain, \emph{i.e.} the large dataset. Afterwards, the pretrained network is finetuned in the target domain, \emph{i.e.} the small cohort, to complete knowledge transfer. We study two transfer learning scenarios with slight and heavy channel mismatch between the source and target domains. We also investigate whether, and if so, how finetuning entirely or partially the pretrained network would affect the performance of sleep staging on the target domain.
		Using the Montreal Archive of Sleep Studies (MASS) database consisting of 200 subjects as the source domain and the Sleep-EDF Expanded database consisting of 20 subjects as the target domain in this study, our experimental results show significant performance improvement on sleep staging achieved with the proposed deep transfer learning approach. Furthermore, these results also reveal the essential of finetuning the feature-learning parts of the pretrained network to be able to bypass the channel mismatch problem.

	\end{abstract}
	
	
	\section{Introduction}
	\label{sec:introduction}
	
	Deep learning has been successfully applied to numerous domains and has received much attention from the sleep research community. With its ability in automatic representation learning to make sense of large datasets, deep learning has improved the performance of automatic sleep staging significantly, reaching an accuracy rate on par with manual scoring by sleep experts \cite{Phan2019a, Stephansen2018}. However, this expert-level performance is only obtainable when the studied cohort is of large size, \emph{i.e.} hundreds to thousands of subjects \cite{Phan2019a, Stephansen2018}. The reason for this is that deep neural networks usually require a large amount of data to train. In practice, many sleep studies possess a small cohort, such as a few dozens of subjects \cite{Kemp2000, Goldberger2000, Olesen2016} sometimes even less \cite{Mikkelsen2018b, Cooray2018}, particularly those studies related to sleep disorders \cite{Cooray2018,Andreotti2018} or exploring new channel positions \cite{Mikkelsen2018b}.
	As a consequence, the small amount of data makes deep neural networks underperform in these sleep studies. One possibility to remedy the lack of data is to make use of the large databases which are manually annotated and publicly available. However, these databases cannot be simply added into these studies due to channel mismatch. 
	The problem of channel mismatch arises when different studies uses different channel layouts \cite{Stephansen2018} or when novel electrode placements might be explored \cite{Mikkelsen2018b}. Moreover, it also happens when a study investigates a particular sleep abnormality, poor performance can be obtained when the automated diagnostic procedure is only trained on healthy volunteers \cite{Cooray2018}. 
	
	We propose a transfer learning approach to circumvent the channel-mismatch problem, enabling efficient knowledge transfer from a large database to study single-channel sleep staging in a much smaller dataset with deep neural networks. We use Montreal Archive of Sleep Studies (MASS) database with 200 subjects \cite{Oreilly2014} as the source domain and the Sleep-EDF Expanded database with 20 subjects \cite{Kemp2000, Goldberger2000} as the target domain in this study. \emph{SeqSleepNet}, a \emph{sequence-to-sequence} model which was recently shown to achieve state-of-the-art results on sleep staging in our recent work \cite{Phan2019a}, is employed as the base model. The network is firstly trained with the source domain data and is then finetuned with the target domain data. We study two scenarios: (1) same-modality transfer learning with slight channel mismatch when the source domain is an EEG channel and the target domain is another EEG channel, (2) cross-modality transfer learning with heavy channel mismatch when the source domain is an EEG channel and the target domain is an EOG channel. Furthermore, we investigate different finetuning strategies to gain insight into deep transfer learning for sleep staging under different channel mismatch conditions.
	
	\section{Sleep Staging Databases}
	\subsection{Montreal Archive of Sleep Studies (MASS)}
	We employed the public MASS dataset \cite{Oreilly2014} as the source domain in this study. This dataset consists of whole-night recordings from 200 subjects aged between 18 and 76 years (97 males and 103 females). Manual annotation was done on each epoch of the recordings by sleep experts according to the AASM standard \cite{Iber2007} (SS1 and SS3 subsets) or the R\&K standard \cite{Hobson1969}  (SS2, SS4, and SS5 subsets). As in \cite{Phan2018e,Phan2019a}, we converted different annotations into five sleep stages \{W, N1, N2, N3, and REM\} and expanded 20-second epochs into 30-second ones by including 5-second segments before and after each epoch. We adopted the C4-A1 EEG channel (C4-A1) as the source domain data. The signals, originally sampled at 256 Hz, were downsampled to 100 Hz.
	
	\subsection{Sleep-EDF Expanded (Sleep-EDF)}
	We adopted the Sleep Cassette (SC) subset of the Sleep-EDF dataset \cite{Kemp2000, Goldberger2000}  as the target domain in this study. It consists of 20 subjects aged 25-34. The PSG recordings, sampled at 100 Hz, of two subsequent day-night periods were available for each subject, except for one subject (subject 13) who had only one-night data. Each 30-second epoch of the recordings was manually labelled by sleep experts according to the R\&K standard \cite{Hobson1969} into one of eight categories \{W, N1, N2, N3, N4, REM, MOVEMENT, UNKNOWN\}. Similar to previous works \cite{Tsinalis2016, Tsinalis2016b, Supratak2017}, N3 and N4 stages were merged into a single stage N3. MOVEMENT and UNKNOWN categories were excluded. We explored the Fpz-Cz EEG channel and the EOG (horizontal) channel in our same-modality and cross-modality transfer learning experiments. Only the in-bed parts (from \emph{lights off} time to \emph{lights on} time) of the recordings were included as recommended in \cite{Imtiaz2015,Tsinalis2016, Tsinalis2016b, Phan2018e}.
	
	\section{SeqSleepNet-based Transfer Learning for Single-Channel Automatic Sleep Staging}
	\label{sec:seqsleepnet}
	
	\subsection{Single-channel SeqSleepNet}
	\emph{SeqSleepNet} \cite{Phan2019a} has demonstrated state-of-the-art performance on the MASS dataset with multi-channel input. We adapt it to study single-channel sleep staging here. The network was proposed to deal with sequence-to-sequence sleep staging, \emph{i.e.} mapping a sequence of multiple consecutive epochs to a sequence of sleep stage labels at once \cite{Phan2019a}. Formally, sequence-to-sequence sleep staging is formulated to maximize the conditional probability 
	$p(\mathbf{y}_1, \mathbf{y}_2, \ldots, \mathbf{y}_L \,|\, \mathbf{S}_1, \mathbf{S}_2, \ldots, \mathbf{S}_L)$ where $\mathbf{S}_1, \mathbf{S}_2, \ldots, \mathbf{S}_L$ denote $L$ consecutive epochs in the input sequence and $\mathbf{y}_1, \mathbf{y}_2, \ldots, \mathbf{y}_L$ represent $L$ one-hot encoding vectors of the ground-truth output labels corresponding to the input epochs.
	
	\begin{figure} [!t]
		\centering
		\includegraphics[width=1\linewidth]{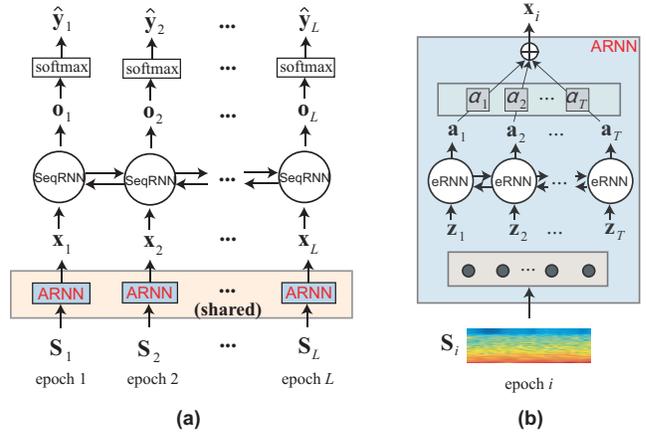}
		\vspace{-1.5cm}
		\caption{(a) \emph{SeqSleepNet}'s overall architecture and (b) its attentional RNN (ARNN) subnetwork.}
		\label{fig:seqsleepnet}
		\vspace{-0.25cm}
	\end{figure}
	
	The overall architecture of \emph{SeqSleepNet} is illustrated in Fig \ref{fig:seqsleepnet}(a). The network is composed of three main components: an attentional RNN (ARNN) subnetwork for epoch-wise feature learning, a sequence-level RNN (SeqRNN) for long-term sequential modelling, and a softmax layer for sequence classification.
	
	{\bf ARNN.} The ARNN subnetwork, illustrated in Fig. \ref{fig:seqsleepnet}(b), is shared between all epochs in the input sequence. It is the combination of a \emph{filterbank} layer \cite{phan2018c}, an epoch-level bidirectional RNN (eRNN for short to distinguish it from the SeqRNN), and an attention layer \cite{Luong2015b}. The subnetwork receives a time-frequency image $\mathbf{S} \in \mathbb{R}^{F\times T}$ obtained from an epoch as input (i.e. an epoch $\mathbf{S}_i$ in the input sequence, cf. Fig.~\ref{fig:seqsleepnet}), where $F$ and $T$ denote the frequency and time dimensions, respectively. The input is firstly preprocessed by the filterbank layer of $M$ filters ($M < F$) for frequency-dimension reduction and smoothing \cite{phan2018c}. Afterwards, the resulting image is treated as a sequence of $T$ local feature vectors (\emph{i.e.} image columns) $(\mathbf{z}_1, \mathbf{z}_2, \ldots,\mathbf{z}_T)$, which is then encoded into a sequence of output vectors $(\mathbf{a}_1, \mathbf{a}_2, \ldots,\mathbf{a}_T)$, where
	\begin{align}
	\mathbf{a}_t &= \mathbf{W}_{ha}[\mathbf{h}^{\text{b}}_t \oplus \mathbf{h}^{\text{f}}_t] + \mathbf{b}_{a}, \mbox{~~} 1 \le t \le T.
	\label{eq:rnn_output}
	\end{align}
	In (\ref{eq:rnn_output}), $\mathbf{W}_{ha}$ denotes the weight matrix, $\mathbf{b}_{a}$ denotes the bias, and $\oplus$ denotes vector concatenation. $\mathbf{h}^{\text{b}}_i$  and $\mathbf{h}^{\text{f}}_i$ represent backward and forward hidden state vectors of the backward and forward recurrent layers at time index $t$, respectively, and are computed as
	\begin{align}
	\mathbf{h}^{\text{f}}_t &= \mathcal{H}(\mathbf{x}_t\,, \mathbf{h}^{\text{f}}_{t-1}), \label{eq:rnn_hidden_forward} \\
	\mathbf{h}^{\text{b}}_t &= \mathcal{H}(\mathbf{x}_t\,, \mathbf{h}^{\text{b}}_{t+1}), \mbox{~} 1 \le t \le T.
	\label{eq:rnn_hidden_backward}
	\end{align}
	Here, $\mathcal{H}$ denotes the hidden layer function which is realized by a Gated Recurrent Unit (GRU) cell \cite{Cho2014}. In turn, the attention layer \cite{Luong2015b} learns to weight the output vectors and collectively form the feature vector $\mathbf{x}$:
	\begin{align}
	\mathbf{x} = \sum\nolimits^{T}_{t=1}\alpha_t\mathbf{a}_t.
	\label{eq:attentive_output}
	\end{align}
	Here, $\alpha_t$ is the attention weight at the time index $t$ learned by the attention layer.
	
	{\bf SeqRNN.} Via the shared ARNN, the input sequence $(\mathbf{S}_1, \mathbf{S}_2, \ldots, \mathbf{S}_L)$ has been transformed into a sequence of epoch-wise feature vectors $(\mathbf{x}_1, \mathbf{x}_2, \ldots,\mathbf{x}_L)$. SeqRNN, which is essentially also a GRU-based bidirectional RNN, reads the induced sequence of feature vectors forward and backward to encode their sequential interactions into a sequence of output vectors $(\mathbf{o}_1, \mathbf{o}_2, \ldots,\mathbf{o}_L)$, where
	\begin{align}
	\mathbf{o}_i &= \mathbf{W}_{ho}[\mathbf{\tilde{h}}^{\text{b}}_i \oplus \mathbf{\tilde{h}}^{\text{f}}_i] + \mathbf{b}_{o},
	\label{eq:seqrnn_output} \\
	\mathbf{\tilde{h}}^{\text{f}}_i &= \mathcal{H}(\mathbf{o}_i\,, \mathbf{\tilde{h}}^{\text{f}}_{i-1}), \label{eq:seqrnn_hidden_forward} \\
	\mathbf{\tilde{h}}^{\text{b}}_i &= \mathcal{H}(\mathbf{x}_i\,, \mathbf{\tilde{h}}^{\text{b}}_{i+1}), \mbox{~} 1 \le i \le L.
	\label{eq:seqrnn_hidden_backward}
	\end{align}
	
	{\bf Softmax.} The softmax layer eventually classifies the sequence of output vectors $(\mathbf{o}_1, \mathbf{o}_2, \ldots,\mathbf{o}_L)$ to produce the output sequence of sleep stage probabilities $(\mathbf{\hat{y}}_1, \mathbf{\hat{y}}_2, \ldots,\mathbf{\hat{y}}_L)$. 
	
	The network is trained to minimize the sequence classification loss \cite{Phan2019a} over $N$ training sequences in the training data:
	\begin{align}
	E(\bm{\theta}) = -\frac{1}{L}\sum_{n=1}^{N}\sum_{i=1}^{L} \mathbf{y}_i\log\left(\mathbf{\hat{y}}_i\left(\bm{\theta}\right)\right) + \frac{\lambda}{2}\|\bm{\theta}\|^2_2.
	\label{eq:sequence_loss}
	\end{align}
	Although being hierarchical, the network is trained in an end-to-end manner similar to \cite{Phan2019a}. 
	
	\section{SeqSleepNet-based Transfer Learning}
	
\begin{figure} [!t]
		\centering
		\includegraphics[width=.65\linewidth]{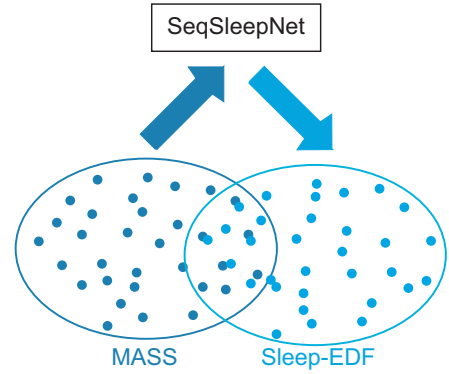}
		\vspace{-0.15cm}
		\caption{\emph{SeqSleepNet}-based transfer learning from the source domain (MASS) to the target domain (Sleep-EDF) with channel mismatch.}
		\label{fig:sleep_transfer}
		\vspace{-0.3cm}
	\end{figure}

	Insufficient training data is a common problem in many sleep studies as it is expensive and difficult to collect a large-scale and high-quality annotated sleep dataset. Meanwhile, there exist large sleep databases which are publicly available, consisting data from hundreds to thousands of subjects, such as MASS dataset \cite{Oreilly2014}. Unfortunately, these datasets cannot be simply included into the private studies due to channel mismatch (see Section \ref{sec:introduction}). Transfer learning \cite{Pan2010} relaxes the hypothesis that the training data must be independent and identically distributed (i.i.d.) from the test data, and therefore, holds promise to leverage these large amount of available data to tackle the problem of insufficient training data in the studies with a small number of subjects. We investigate deep transfer learning with \emph{SeqSleepNet} as the base model, the MASS database as the source domain, and the Sleep-EDF database as the target domain. A \emph{SeqSleepNet} will be used as a device to transfer the knowledge from the MASS database for sleep staging with the Sleep-EDF database as illustrated in Fig. \ref{fig:sleep_transfer}.
	
	We study the two transfer learning scenarios for single-channel sleep staging, taking into account different degrees of channel mismatch:
	
	{\bf EEG$\mapsto$EEG:} In this same-modality transfer learning, \emph{SeqSleepNet} is firstly trained on the C4-A1 EEG channel of the MASS database and is then transferred to the Fpz-Cz EEG channel of the Sleep-EDF database. This study is to showcase that even the source domain and the target domain are of the same modality, transfer learning is still necessary to overcome the small channel mismatch and to improve the sleep staging performance. Note that we adopt two different EEG channels for the source and target domains on purpose not only because the C4-A1 EEG channel is not available in the Sleep-EDF database but also because we want to enforce more channel mismatch here.
	
	{\bf EEG$\mapsto$EOG:} In this cross-modality transfer learning, \emph{SeqSleepNet} is firstly trained on the C4-A1 EEG channel of the MASS database and is then transferred to the horizontal EOG channel of the Sleep-EDF database. On the one hand, this study is to demonstrate that transfer learning is essential to tackle heavy channel mismatch between the source and target domains to improve sleep staging performance on the target domain. On the other hand, as EOG signals contain rich information from multiple sources, including ocular activity, frontal EEG activity, and EMG from cranial and eye muscles, they are promising candidates for a single-modality sleep staging system. Despite their potential, EOG signals have been mainly used as a complement for EEG signals in multi-modality sleep staging studies \cite{Phan2019a,Phan2018e}. Only a few studies have exploited standalone EOG signals for single-modality sleep staging \cite{Olesen2016,Liang2015}. We therefore want to examine whether deep transfer learning could help sleep staging with standalone secondary EOG to achieve a performance level comparable to that using the primary EEG. Furthermore, due to the ease of electrode placements, they would be ideal for home-based sleep monitoring with wearable devices \cite{Mikkelsen2018b}. 
	
	In order to accomplish transfer learning, the \emph{SeqSleepNet} pretrained in the source domain is considered as a starting point in the target domain. We reuse the entire or parts of the pretrained network and finetune the rest with the target domain data to study the influence of finetuning different parts of the pretrained network to the sleep stage performance on the target domain. To that end, we examine directly transferring (\emph{i.e.} the pretrained network is directly used in the target domain without finetuning), finetuning the entire pretrained network, and partially finetuning \{softmax, softmax + ARNN, softmax + SeqRNN\} combinations of the pretrained network while other layers stay fixed.
	
	\vspace{-0.1cm}
	\section{Experiments}
	\vspace{-0.1cm}
	\subsection{Experimental setup}
	While the entire 200 subjects of the MASS database (\emph{i.e.} the source domain) was used for network pretraining, we conducted leave-one-subject-out cross validation on the Sleep-EDF database (\emph{i.e.} the target domain) to access the efficiency of transfer learning on the target domain. At each iteration, 19 remaining subjects of Sleep-EDF were further divided into 15 subjects for finetuning the pretrained model and 4 subjects for validation. The performance over 20 cross-validation folds will be reported.
	
	\vspace{-0.1cm}
	\subsection{Time-frequency input}
	As \emph{SeqSleepNet} receives time-frequency images as input, each 30-second epoch (\emph{i.e.} EEG and EOG) was transformed into log-power spectra via short-time Fourier transform (STFT) with a window size of two seconds and 50\% overlap, followed by logarithmic scaling. Hamming window and 256-point Fast Fourier Transform (FFT) were used. This results in an image $\mathbf{S} \in \mathbb{R}^{F \times T}$ where $F = 129$ (the number of frequency bins), $T = 29$ (the number of spectral columns).
	
	\vspace{-0.15cm}
	\subsection{Network parameters}
	The implementation was based on \emph{Tensorflow} framework \cite{Abadi2016}. For pretraining, the network was trained with the MASS dataset for 10 training epochs with a minibatch size of 32 sequences. For finetuning, a pretrained network was further trained with the Sleep-EDF training data for 10 training epochs. Early stopping was exercised, during which the finetuning was stopped if after 50 training steps there was not any improvement on the Sleep-EDF validation set. The network training and finetuning were performed using \emph{Adam} optimizer with a learning rate of $10^{-4}$.
	
	The network was parametrized similarly to that in our previous work \cite{Phan2019a} (i.e. the number of filters $M$, the size of the GRUs' hidden state vectors, dropout, etc.) except that its input is single-channel here. 
	We experimented with sequence length $L=20$ epochs. Similar to \cite{Phan2019a}, the sequences were sampled from the PSG recordings with a maximum overlap (\emph{i.e.} $L-1$ epochs) to generate all possible sequences from the training recordings for network training and finetuning. During testing, the test sequence was shifted by one to obtain an ensemble of classification decisions at each epoch of the test PSG recordings. Probabilistic multiplicative fusion was then carried out to aggregate these decisions to produce the final decision as in \cite{Phan2019a}.

	\vspace{-0.15cm}
	\subsection{Experimental results}
	\subsubsection{Results on {\bf EEG$\mapsto$EEG} transfer learning}
	Table \ref{tab:performance} shows the performance of the {\bf EEG$\mapsto$EEG} transfer learning scenario. We also compare these results with those reported in the literature. The overall performance is reported using accuracy, macro F1-score (MF1), Cohen's kappa ($\kappa$), sensitivity, and specificity. Note that, as Sleep-EDF was used differently in previous works \cite{Phan2018e, Imtiaz2015}, for compatible comparison, we only included those with a similar experimental setting (\emph{i.e.} conducting independent testing with single FPz-Cz channel and in-bed data only \cite{Phan2018e}). A more comprehensive comparison can be found in \cite{Phan2018e}. 
	
	The results in Table \ref{tab:performance} show that, under the condition of small channel mismatch, direct transfer obtains a good accuracy level which is close to the best reported in previous work (\emph{i.e.} Multitask 1-max CNN \cite{Phan2018e}). However, this accuracy is $1.0\%$ lower than the \emph{SeqSleepNet} trained from scratch with the target domain data only, highlighting the necessity of finetuning even with same-modality transfer learning. As expected, entire or partial finetuning the pretrained \emph{SeqSleepNet} results in significant performance gain over the naive direct transfer strategy. Absolute gains of $0.4\%$, $4.0\%$, $4.2\%$, and $4.3\%$ on overall accuracy are obtained when finetuning softmax, softmax + ARNN, softmax + SeqRNN, and all layers of the pretrained network, respectively. Moreover, the finetuning results significantly outperform the best result obtained in previous work, improving the single-channel sleep staging accuracy on the Sleep-EDF from $81.9\%$ (\emph{i.e.} Multitask 1-max CNN \cite{Phan2018e}) to $85.5\%$ (\emph{i.e.} finetuning entire network). 
	
\begin{figure*} [!t]
		\centering
		\includegraphics[width=0.825\linewidth]{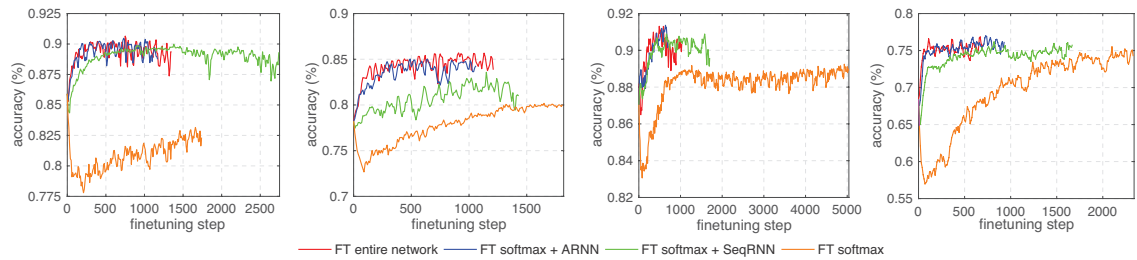}
		\vspace{-0.3cm}
		\caption{Variation of the test accuracy as a function of the finetuning step.}
		\label{fig:tuning_curve}
		\vspace{-0.35cm}
	\end{figure*}

	The finetuning results also unveil that finetuning the softmax layer alone is not sufficient to overcome the channel-mismatch obstacle. Instead, it is important to additionally finetune those feature-learning layers, either the ARNN subnetwork for epoch-level feature learning or the SeqRNN for sequence-level feature learning or both collectively. In Fig. \ref{fig:tuning_curve}, we show the development curves of the test accuracy of four different cross-validation folds during finetuning. Earlier convergence and better generalization can be consistently seen when finetuning the feature-learning parts of the network in addition to the softmax layer.
	
	\setlength\tabcolsep{2.25pt}
	\begin{table}[!t]
		\caption{Results of the {\bf EEG$\mapsto$EEG} transfer learning in comparison with results reported in previous work. Finetuning is abbreviated as FT for short.}
		\vspace{-0.3cm}
		\footnotesize
		\begin{center}
			\begin{tabular}{|>{\arraybackslash}m{1.25in}|>{\centering\arraybackslash}m{0.4in}|>{\centering\arraybackslash}m{0.25in}|>{\centering\arraybackslash}m{0.3in}|>{\centering\arraybackslash}m{0.25in}|>{\centering\arraybackslash}m{0.25in}|>{\centering\arraybackslash}m{0.25in}|>{\centering\arraybackslash}m{0in} @{}m{0pt}@{}}
				\cline{1-7}
				\multirow{2}{*}{System} & \multirow{2}{*}{\makecell{Transfer \\learning}} & \multicolumn{5}{c|}{Overall metrics} & \parbox{0pt}{\rule{0pt}{0.25ex+\baselineskip}} \\ [0ex]  	
				
				\cline{3-7}
				& & Acc. & $\kappa$ & MF1 & Sens. & Spec. & \parbox{0pt}{\rule{0pt}{0.25ex+\baselineskip}} \\ [0ex]  	
				\cline{1-7}
				
				FT Entire network & Yes & $\bm{85.5}$ & $\bm{0.794}$ & $\bm{80.0}$ & $\bm{79.6}$ & $\bm{95.9}$ & \parbox{0pt}{\rule{0pt}{0.25ex+\baselineskip}} \\ [0ex]  	
				
				FT Softmax + ARNN & Yes & $85.4$ & $0.792$ & $79.9$ & $79.6$ & $95.9$ &  \parbox{0pt}{\rule{0pt}{0.25ex+\baselineskip}} \\ [0ex]  	
				
				FT Softmax + SeqRNN & Yes & $85.2$ & $0.789$ & $79.6$ & $78.7$ & $95.8$ &   \parbox{0pt}{\rule{0pt}{0.25ex+\baselineskip}} \\ [0ex]  	
				
				FT Softmax & Yes & $81.6$ & $0.735$ & $74.9$ & $74.4$ & $94.8$ & \parbox{0pt}{\rule{0pt}{0.25ex+\baselineskip}} \\ [0ex]  	
				
				Direct transfer (no FT) & Yes & $81.2$ & $0.733$ & $74.6$ & $74.1$ & $94.8$ &  \parbox{0pt}{\rule{0pt}{0.25ex+\baselineskip}} \\ [0ex]  	
				
				\cline{1-7}	
				
				\emph{SeqSleepNet} (scratch)  & No & $82.2$ & $0.746$ & $74.1$ & $73.9$ & $95.0$ &  \parbox{0pt}{\rule{0pt}{0.25ex+\baselineskip}} \\ [0ex]  	
				
				Multitask 1-max CNN \cite{Phan2018e} & No & $81.9$ & $0.740$ & $73.8$ & $73.9$ & $95.0$ &  \parbox{0pt}{\rule{0pt}{0.25ex+\baselineskip}} \\ [0ex]  	
				
				1-max CNN \cite{phan2018c} & No & $79.8$ & $0.720$ & $72.0$ & $-$ & $-$ &  \parbox{0pt}{\rule{0pt}{0.25ex+\baselineskip}} \\ [0ex]  	
				
				Atentional RNN \cite{phan2018d} & No & $79.1$ & $0.700$ & $69.8$ & $-$ & $-$ &  \parbox{0pt}{\rule{0pt}{0.25ex+\baselineskip}} \\ [0ex]  	
				
				Deep auto-encoder \cite{Tsinalis2016b} & No & $78.9$ & $-$ & $73.3$ & $-$ & $-$ &  \parbox{0pt}{\rule{0pt}{0.25ex+\baselineskip}} \\ [0ex]  	
				
				Deep CNN \cite{Tsinalis2016} & No & $74.8$ & $-$ & $69.8$ & $-$ & $-$ &  \parbox{0pt}{\rule{0pt}{0.25ex+\baselineskip}} \\ [0ex]  	
				\cline{1-7}
			\end{tabular}
		\end{center}
		\label{tab:performance}
		\vspace{-0.35cm}
	\end{table}
	
	\setlength\tabcolsep{2.25pt}
	\begin{table}[!t]
		\caption{Results of the {\bf EEG$\mapsto$EOG} transfer learning. Finetuning is abbreviated as FT for short.}
		\vspace{-0.3cm}
		\footnotesize
		\begin{center}
			\begin{tabular}{|>{\arraybackslash}m{1.25in}|>{\centering\arraybackslash}m{0.4in}|>{\centering\arraybackslash}m{0.25in}|>{\centering\arraybackslash}m{0.3in}|>{\centering\arraybackslash}m{0.25in}|>{\centering\arraybackslash}m{0.25in}|>{\centering\arraybackslash}m{0.25in}|>{\centering\arraybackslash}m{0in} @{}m{0pt}@{}}
				\cline{1-7}
				\multirow{2}{*}{System} & \multirow{2}{*}{\makecell{Transfer \\learning}} & \multicolumn{5}{c|}{Overall metrics} & \parbox{0pt}{\rule{0pt}{0.25ex+\baselineskip}} \\ [0ex]  	
				
				\cline{3-7}
				 &  & Acc. & $\kappa$ & MF1 & Sens. & Spec. & \parbox{0pt}{\rule{0pt}{0.25ex+\baselineskip}} \\ [0ex]  	
				\cline{1-7}
				
				FT Entire Network & Yes & $80.8$ & $0.725$ & $74.2$ & $73.1$ & $94.5$ &  \parbox{0pt}{\rule{0pt}{0.25ex+\baselineskip}} \\ [0ex]  	
				
				FT Softmax + ARNN & Yes & $80.1$ & $0.716$ & $73.5$ & $72.5$ & $94.3$ &   \parbox{0pt}{\rule{0pt}{0.25ex+\baselineskip}} \\ [0ex]  	
				
				FT Softmax + SeqRNN & Yes & $80.0$ & $0.709$ & $72.3$ & $70.4$ & $94.0$ &  \parbox{0pt}{\rule{0pt}{0.25ex+\baselineskip}} \\ [0ex]  	
				
				FT Softmax & Yes & $73.2$ & $0.605$ & $62.5$ & $60.7$ & $91.9$ &  \parbox{0pt}{\rule{0pt}{0.25ex+\baselineskip}} \\ [0ex]  	
				
				Direct transfer (no FT) & Yes & $51.1$ & $0.300$ & $42.5$ & $42.9$ & $85.9$ &   \parbox{0pt}{\rule{0pt}{0.25ex+\baselineskip}} \\ [0ex]  	
				
				\cline{1-7}	
				
				\emph{SeqSleepNet} (scratch)  & No & $78.5$ & $0.688$ & $68.3$ & $67.4$ & $93.6$ &   \parbox{0pt}{\rule{0pt}{0.25ex+\baselineskip}} \\ [0ex]  	
				
				\cline{1-7}
			\end{tabular}
		\end{center}
		\label{tab:performance_eog}
		\vspace{-0.35cm}
	\end{table}

	\subsubsection{Results on {\bf EEG$\mapsto$EOG} transfer learning}
	Table \ref{tab:performance_eog} shows the results of the {\bf EEG$\mapsto$EOG} transfer learning in comparison to that obtained by the \emph{SeqSleepNet} trained from scratch using the target domain data only. Firstly, it can be seen that direct transfer does not work properly under the heavy channel-mismatch condition. It is understandable that differences in characteristics of the source EEG and the target EOG cause malfunction of the feature-learning parts of the pretrained \emph{SeqSleepNet} in the target domain. As a consequence, finetuning is essential in this case. Secondly, the finetuning results exhibit similar patterns to the same-modality transfer learning case. That is, it is chief to finetune the feature-learning layers, either the ARNN subnetwork or the SeqRNN or both, alongside with the softmax layer. Finetuning leads to absolute gains of $22.1\%$, $28.9\%$, $29.0\%$, and $29.7\%$ on overall accuracy compared  to the naive direct transfer when softmax, softmax + ARNN, softmax + SeqRNN, and all layers of the pretrained network are finetuned, respectively.
	
	More importantly, despite the heavy channel mismatch condition, transferring the knowledge of the source domain to the target domain brings up the accuracy by $2.3\%$ over the network trained from scratch with the target domain data only. With the obtained accuracy of $80.8\%$, single-channel sleep staging with the secondary EOG reaches, and sometimes even outperforms, many deep-learning approaches based on the primary EEG presented in Table \ref{tab:performance}. Therefore, this finding would allow one to explore the usage of EOG as an alternative for EEG in many applications, such as home-based sleep monitoring with wearable devices \cite{Mikkelsen2018b,Looney2016}. 
	
	\vspace{-0.15cm}
	\section{Conclusions}
	We propose a deep transfer learning approach using the state-of-the-art \emph{SeqSleepNet} to borrow and transfer knowledge from the large MASS database (the source domain) to enable exploiting the modelling power of deep neural networks on single-channel sleep staging with the Sleep-EDF database (\emph{i.e.} the target domain) with a small number of subjects. \emph{SeqSleepNet} was firstly trained with the C4-A1 EEG channel of the MASS database and was then transferred to the Fpz-Cz EEG channel and the horizontal EOG channel of the Sleep-EDF database via finetuning. The experimental results showed that the sleep staging performance on the target domain was significantly improved in both same-modality and cross-modality transfer learning scenarios. In addition, we demonstrated that it is essential to finetune the feature-learning parts of the pretrained network to overcome the channel mismatch problem.
	
	\vspace{-0.15cm}
	\bibliographystyle{IEEEbib}
	\bibliography{bibliography}
	
\end{document}